\documentclass[letterpaper]{article} 
\usepackage[submission]{aaai23}  
\usepackage{times}  
\usepackage{helvet}  
\usepackage{courier}  
\usepackage[hyphens]{url}  
\usepackage{graphicx} 
\usepackage{natbib}  
\usepackage{caption} 
\usepackage{algorithm}
\usepackage{algorithmic}
\usepackage{comment}
\usepackage{cite}

\usepackage{newfloat}
\usepackage{listings}
\DeclareCaptionStyle{ruled}{labelfont=normalfont,labelsep=colon,strut=off} 
\floatstyle{ruled}
\newfloat{listing}{tb}{lst}{}
\floatname{listing}{Listing}

\usepackage{xspace}
\usepackage{multirow}
\newcommand{\logistics}{\textsc{Logistics}\xspace}
\newcommand{\blocksworld}{\textsc{Blocksworld}\xspace}
\newcommand{\satellite}{\ensuremath{\textsc{Satellite}}\xspace}
\newcommand{\zenotravel}{\textsc{Zenotravel}\xspace}
\newcommand{\driverlog}{\textsc{Driverlog}\xspace}
\newcommand{\depots}{\textsc{Depots}\xspace}
\newcommand{\lpg}{\textsc{lpg}\xspace}
\newcommand{\lama}{\textsc{lama}\xspace}
\newcommand{\set}[1]{\ensuremath{\{#1\}}\xspace}

\newcommand{\pereiraGR}{{\ensuremath{\sf LGR}}\xspace}
\newcommand{\our}{\ensuremath{{\sf GRNet}}\xspace}

\newcommand{\pddl}{\textsc{pddl}\xspace}

\newcommand{\obs}{\textit{obs}}

\newcommand{\pereiraTEST}{\ensuremath{\textsc{TS}_{\textit{Per}}}\xspace}
\newcommand{\ourTEST}{\ensuremath{\textsc{TS}_{\textit{PerGen}}}\xspace}
\newcommand{\recogTEST}{\ensuremath{\textsc{TS}_{\textit{Rec}}}\xspace}

\newtheorem{example}{Example}

\usepackage{xcolor}

\usepackage{mathtools}
\DeclarePairedDelimiter\abs{\lvert}{\rvert}%

\usepackage{booktabs}
\usepackage{subcaption}
\usepackage{graphicx}
\usepackage{amsfonts}
\usepackage{mathtools}

\title{Goal Recognition as a Deep Learning Task: the GRNet Approach}
\author{
    Mattia Chiari,\textsuperscript{\rm 1}
    Alfonso E. Gerevini,\textsuperscript{\rm 1}
    Luca Putelli,\textsuperscript{\rm 1}
    Francesco Percassi,\textsuperscript{\rm 2}
    Ivan Serina,\textsuperscript{\rm 1}
}
\affiliations{
    \textsuperscript{\rm 1}Dipartimento di Ingegneria dell'Informazione, Università degli Studi di Brescia, Via Branze 38, Brescia, Italy\\
    \textsuperscript{\rm 2}School of Computing and Engineering, University of Huddersfield, Queensgate, Huddersfield HD1 3DH, United Kingdom\\
    \{m.chiari017, alfonso.gerevini, luca.putelli1, ivan.serina\}@unibs.it\\
    f.percassi@hud.ac.uk
}

\begin{document}

\maketitle

\begin{abstract}
In automated planning, recognising the goal of an agent from a trace of observations is an important task with many applications. The state-of-the-art approaches to goal recognition
rely on the application of planning techniques, which requires a model of the domain actions and of the initial domain state (written, e.g., in \pddl). 
We study an alternative approach where goal recognition is formulated as a classification task addressed by machine learning.
Our approach, called \our, is primarily aimed at making goal recognition more accurate as well as faster by learning how to solve it in a given domain.
Given a planning domain specified by a set of propositions and a set of action names, the goal classification instances in the domain are solved by a Recurrent Neural Network (RNN). A run of the RNN processes a trace of observed actions to compute how likely it is that each domain proposition is part of the agent's goal, for the problem instance under consideration. These predictions are then aggregated to choose one of the candidate goals. 
The only information required as input of the trained RNN is a trace of action labels,  
each one indicating just the name of an observed action. 
An experimental analysis confirms that \our achieves good performance in terms of both goal classification accuracy and runtime, obtaining better performance w.r.t. a state-of-the-art goal recognition system over the considered benchmarks.
\end{abstract}

\section{Introduction}
Goal Recognition is the task of recognising the goal that an agent is trying to achieve from observations about the agent's behaviour in the environment \cite{DBLP:journals/firai/Van-HorenbekeP21,2018_geff_modelfree_modelbased}.
Typically, such observations consist of a trace (sequence) of executed actions in an agent's plan to achieve the goal, or a trace of world states generated by the agent's actions, while an agent goal is specified by a set of propositions.
Goal recognition has been studied in AI for many years, and it is an important task in several fields, including human-computer interactions \cite{human_interaction}, computer games \cite{min}, network security \cite{mirsky2019new}, smart homes \cite{harman2019action}, financial applications \cite{borrajo}, and others.

In the literature, several systems to solve goal recognition problems have been proposed \cite{DBLP:conf/ijcai/MeneguzziP21}. 
The state-of-the-art approach is based on transforming a plan recognition problem into one or more plan generation problems solved by classical planning algorithms \cite{ramirez2009plan,pereira2020landmark,gr_noisy}. 
In order to perform planning, this approach requires domain knowledge consisting of a model of each agent's action specified as a set of preconditions and effects, and a description of an initial state of the world, in which the agent perform the actions. 
The computational efficiency (runtime) largely depends on the planning algorithm performance, which could be inadequate in a context demanding fast goal recognition (e.g., in real-time/online applications).\footnote{
Deciding plan existence in classical planning is PSPACE-complete  \cite{BYLANDER1994}.} 

In this paper, we investigate an alternative approach
in which the goal recognition problem is formulated as a classification task, addressed through machine learning, where each candidate goal (a set of propositions) of the problem can be seen as a value class.
The primary aim is making goal recognition more accurate as well as faster by learning how to solve it in a given domain.
Given a planning domain specified by a set of propositions and a set of action names, we tackle the goal classification instances in the domain through a Recurrent Neural Network (RNN). A run of our RNN processes a trace of observed actions to compute how likely it is that each domain proposition is part of the agent's goal, for the problem instance under considerations. These predictions are then aggregated through a goal selection mechanism   to choose one of the candidate goals.

The proposed approach, that we call {\sf \small GRNet}, is generally faster than the model-based approach to goal recognition based on planning, since running a trained neural network can be much faster than plan generation. Moreover, {\sf \small GRNet}
 operates with minimal information, since the only information required as input for the trained RNN is a trace of action labels (each one indicating just the name of an observed action), and the initial state can be completely unknown.

The RNN is trained only once for a given domain, i.e., the same trained network can be used to solve a large set of goal recognition instances in the domain. On the other hand, as usual in supervised learning,  a  (possibly large) dataset of solved goal recognition instances for the domain under consideration is needed for the training.
When such data are unavailable or scarse, they can be synthetized via planning.
In such a case, the resulting overall system can be seen as a combined approach (model-based for generating training data, and model-free for the goal classification task) that outperforms the pure model-based approach in terms of both classification accuracy and classification runtime.
Indeed, an experimental analysis presented in the paper confirms that \our achieves good performance in terms of both goal classification accuracy and runtime, obtaining consistently better performance with respect to a state-of-the-art goal recognition system over a class of benchmarks in six planning domains.

In rest of paper, after giving background and preliminaries about goal recognition and LSTM networks, we describe the \our approach; then we present the experimental results; finally, we discuss related work and give the conclusions.

\section{Preliminaries}

We describe the problem goal of recognition, starting with its relation to activity/plan recognition, and we give the essential background on Long Short-Term Memory networks and the attention mechanism.

\label{sec:preliminaries}

\subsection{Activity, Goal and Plan Recognition}

Activity, plan, and goal recognition are related tasks \cite{geib2007plan}. Since 
in the literature sometime they are not clearly distinguished, we begin with an informal definition of them following \cite{DBLP:journals/firai/Van-HorenbekeP21}.

Activity recognition concerns analyzing temporal sequences of (typically low-level) data generated by humans, or other autonomous agents acting in an environment, to identify the corresponding activity that they are performing.
For instance, data can be collected from wearable sensors, accelerometers, or images to recognize human activities such as running, cooking, driving, etc. \cite{DBLP:journals/firai/VrigkasNK15, JOBANPUTRA2019698}.

Goal recognition (GR) can be defined as the problem of identifying the intention (goal) of an agent from observations about the agent behaviour in an environment. These observations can be represented as an ordered sequence of discrete actions (each one possibly identified by activity recognition), while the agent's goal can be expressed either as a set of propositions or a probability distribution over alternative sets of propositions (each oven forming a distinct candidate goal).

Finally, 
plan recognition is more general than GR and concerns both recognising the goal of an agent and identifying the full ordered set of actions (plan) that have been, or will be, performed by the agent in order to reach that goal; as GR, typically plan recognition takes as input a set of observed actions performed by the agent \cite{DBLP:journals/umuai/Carberry01}.

\subsection{Model-based and Model-free Goal Recognition}

In the approach to GR known as ``{goal recognition over a domain theory}'' \cite{geffner,DBLP:journals/firai/Van-HorenbekeP21,santos,gr_noisy}, 
the available knowledge consists of an underlying model of the behaviour of the agent and its environment. Such a model 
represents the agent/environment states and the set of actions $A$ that the agent can perform; typically it is specified by a planning language such as 
\pddl \cite{mcdermott1998pddl}.
The states of the agent and environment are formalised as subsets of a set of propositions $F$, called {\em fluents} or {\em facts}, and each domain action in $A$ is modeled by a set of preconditions and a set of effects, both over $F$.  An instance of the GR problem in a given domain is then specified by: 
\begin{itemize}
\item
an initial state $I$ of the agent and environment ($I \subseteq F$);
\item a sequence $O = \langle \obs_1, .., \obs_n \rangle$  of observations ($n \geq 1$), where each $obs_i$ is an action in $A$ performed by the agent;
\item  and a set $\mathcal{G} = \{ G_1,  .., G_m \}$ ($m \geq 1$) of possible goals of the agent, where each $G_i$ is a set of fluents over $F$. 
\end{itemize}
The observations form a trace of the full sequence $\pi$ of actions performed by the agent to achieve the goal.
Such a plan trace is a selection of (possibly non-consecutive)  actions in $\pi$, ordered as in $\pi$.
Solving a GR instance consists of identifying the $G^*$ in $\cal G$ that is the (hidden) goal of the agent. 

The approach based on a model of the agent's actions and of the agent/environment states, that we call {\em model-based goal recognition} (MBGR), defines GR as a reasoning task addressable by automated planning techniques \cite{DBLP:conf/ijcai/MeneguzziP21,GNT2016}. 

An alternative approach to MBGR is {\em model-free goal recognition} (MFGR)  \cite{2018_geff_modelfree_modelbased,borrajo}. In this approach, GR 
is formulated as a classification task addressed through machine learning.
The domain specification consists of a fluent set $F$, and a set of possible actions $A$, where each action $a\in A$ is specified by just a label (a unique identifier for each action).

A MFGR instance for a domain is specified by an observation sequence $O$
formed by action labels and, as in MBGR, a goal set $\cal G$ formed by subsets of $F$.
MFGR requires minimal information about the domain actions, and can operate without the specification of an initial state, that can be completely unknown. 
Moreover, since running a learned classification model is usually fast, a MFGR system is expected to run faster than a MBGR system based on planning algorithms.
On the other hand, MFGR needs a data set of solved GR instances from which learning a classification model for the new GR instances of the domain.

\begin{example}

As a running example, we will use a very simple GR instance in the well-known \blocksworld domain. In this domain the agent has the goal of building one or more stacks of blocks, and only one block may be moved at a time. The agent can perform four types of actions:
{\tt \small Pick-Up} a block from the table, {\tt \small Put-Down} a block on the table,
 {\tt \small Stack} a block on top of another one, and {\tt \small Unstack} a block that is on another one.  
We assume that a GR instance in the domain involves at most 22 blocks. In \blocksworld
there are three types of facts (predicates):
{\small \tt On}, that has two blocks as arguments, plus {\small \tt On-Table} and {\small \tt Clear} that have one argument. Therefore, the fluent set $F$ consists of $22\times 21 + 22 + 22 = 506$ propositions.  
The goal set ${\cal G}$ of the instance example consists of the two goals
$G_1=$ \{{\tt \small(On Block\_F Block\_C)}, {\tt \small (On Block\_C Block\_B)}\} and  $G_2=$ \{{\tt \small (On Block\_G Block\_H)}, {\tt \small (On Block\_H Block\_F)}\}; the observation sequence $O$ is $\langle${\tt \small (Pick-Up Block\_C)}, {\tt \small (Stack Block\_C Block\_B)}, {\tt \small (Pick-Up Block\_F)}$\rangle$. 
\end{example}
\begin{figure*}
    \centering
    \includegraphics[scale=0.28]{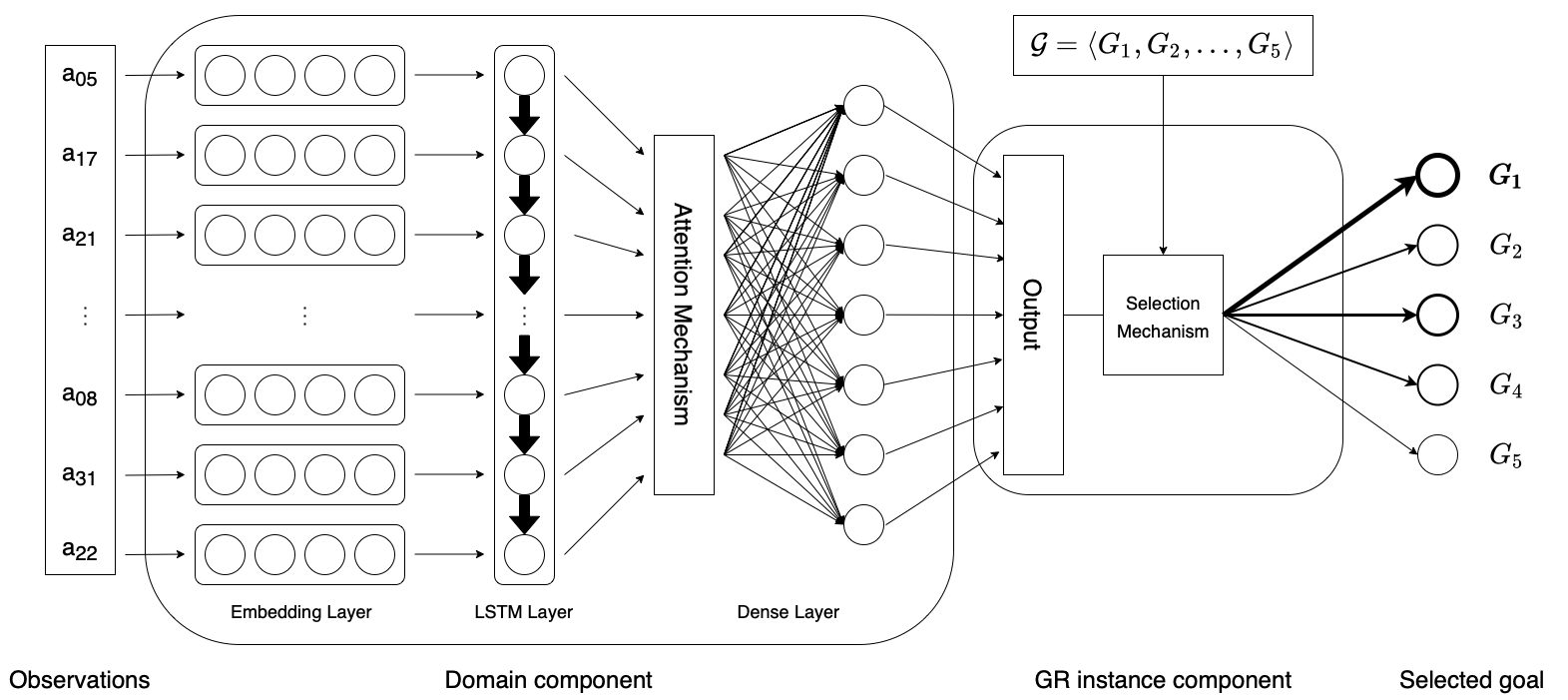}
    \caption{
    Architecture of \our. The input observations are encoded by embedding vectors and then fed to a LSTM neural network. After that the attention mechanism computes the context vector, which is used by a feed-forward layer to define the corresponding output values. This layer is composed by $|F|$ neurons, each one representing a possible fluent in the domain.
    The output of the neural network is then used by the instance component for selecting the goal with the highest score ($G_1$ in the example of the figure).
    The observed actions $a_{05}$, $a_{17}$, ..., $a_{22}$ are ordered from top to bottom according to their execution order.
    }
    \label{fig:architecture}
\end{figure*}

\subsection{LSTM and Attention Mechanism}
\label{sec:lstm}
A Long Short-Term Memory Network (LSTM) is a particular kind of Recurrent Neural Network (RNN). This kind of deep learning architecture is particularly suitable for processing sequential data like signals or text documents \cite{lstm}. With respect to the standard RNN, LSTM deals with typical issues such as vanishing gradient and long-term dependencies, obtaining better predictive performance \cite{Gers2000LearningTF}. Let $x_1, x_2 \ldots x_m$ be an input time series where $x_t \in \mathbb{R}^d$ is the feature vector representing the $t$-th element of the series, and $d$ is the dimension of each feature vector of the sequence. The long and short term memory states $c_{t}\in \mathbb{R}^N$  and $h_{t}\in \mathbb{R}^N$  at time step $t$ of the series, respectively, are computed recursively considering the values at the previous time step $t-1$  as follows:

\hspace{-0.2cm}
{\small
\begin{tabular}{ll}
$\hat{c}_t = \tanh(W_c [h_{t-1}, x_t] + b_c)$ &
$i_t = \sigma(W_i [h_{t-1}, x_t]  + b_i)$ \\
$f_t = \sigma(W_f [h_{t-1}, x_t]  + b_f)$&
$c_t = i_t * \hat{c}_t + f_t * c_{t-1} $\\
$o_t = \sigma(W_o [h_{t-1}, x_t] + b_o)$&
$h_t = \tanh(c_t) * o_t$\\
\end{tabular}
}
\noindent
where $\sigma$ denotes the sigmoid activation function and $*$ corresponds to the element-wise product; $W_f$, $W_i$, $W_o$, $W_c$ $\in \mathbb{R}^{(N+d)\times{N}}$ are the weight matrices and  $b_f$, $b_i$, $b_o$, $b_c$ $\in \mathbb{R}^N$ are the bias vectors; the vectors in square brackets are concatenated.
Weight matrices and bias vectors are typically initialized with the Glorot uniform initializer \cite{glorot2010understanding}, and they are shared by all the cells in the LSTM layer. $h_0$ and $c_0$ are initialized as zero vectors.

The {\it attention mechanism} \cite{DBLP:journals/corr/BahdanauCB14} is another layer which computes weights representing the contribution of each element of the sequence, and provides a 
representation of the sequence (also called the \textit{context vector}) as the weighted average of the outputs ($h_t$) of the LSTM cells, 
improving the predictive performance with respect to the base LSTM networks.
In our system, we use the so-called \textit{word attention} introduced by \citet{yang_attention} in the context of text classification.

\section{Goal Recognition through GRNet}

In this section we present our approach to goal recognition based on deep learning, \our. 
\our is depicted in Figure \ref{fig:architecture}  consists of two main components.
The first component takes as input the observations of the GR instance to solve, and gives as output a score (between 0 and 1) for each proposition in the domain proposition set $F$. This component,  called {\it Domain Component}, is general in the sense that it can be used for every GR instance over $F$ (training is performed once for each domain). 
The second component, called {\em Instance Component}, takes as input the proposition ranks generated by the domain component for a GR instance, and uses them to select a goal from the candidate goal set ${\cal G}$.

\subsection{The Domain Component of GRNet}
Given a sequence of observations, represented on  the left side of Figure \ref{fig:architecture}, each action $a_i$ corresponding to an observation is encoded as a vector $e_i$ of real numbers by an embedding layer \cite{bengio2003neural}.\footnote{\url{https://keras.io/api/layers/core_layers/embedding/}}
In Figure \ref{fig:architecture}, the observed actions are displayed from top to bottom in the order in which they are executed by the agent. The embedding layer is initialised with random weights, and trained at the same time with the rest of the domain component.

The index of each observed action is simply the result of an arbitrary order of the actions that is computed in the pre-processing phase, only once for the domain under consideration.
Please note that two actions $a_i$ and $a_j$ consecutively observed may not be consecutive actions in the full plan of the agent
(the full plan may contain any number of actions between $a_i$ and $a_j$).

The sequence of embedding vectors is then fed to a LSTM Neural Network, and the result of the output of each cell is processed by the Attention Mechanism.
After computing a weight for the contribution of each cell, this layer provides a so-called \textit{context-vector} 
that summarizes the information contained in the trace of plan. 
The context vector is then passed to a feed-forward layer, which has $N$ output neurons with \textit{sigmoid} activation function. 
$N$ is the number of the domain fluents (propositions) that can appear in any goal of {$\cal G$} for any GR instance in the domain; for our experiments $N$ was set to the size of the domain fluent set $F$, i.e., $ N = |F|$.
The output of the $i$-th neuron $\overline{o}_i$ corresponds to the $i$-th fluent $f_i$ (fluents are lexically ordered), and the activation value of $\overline{o}_i$ gives a rank for $f_i$ being true in the agent's goal (with rank equal to one meaning that $f_i$ is true in the goal).
In other words, our network is trained as a multi-label classification problem, where each domain fluent can be considered as a different binary class. As loss function, we used standard binary crossentropy.

As shown in Figure \ref{fig:architecture}, the dimension of the input and output of our neural networks depend only on the selected domain and some basic information, such as the maximum number of possible output facts
that we want to consider.
The dimension of the embedding vectors, the dimension of the LSTM layer and other hyperaparameters of the networks are selected using the Bayesian-optimisation approach provided by the Optuna framework \cite{akiba2019optuna}, with a validation set formed by 20\% of the training set while the remaining 80\% is used for training the network.
More details about the hyperparameters are given in the Supplementary Material. 

\subsection{The Instance Specific Component of GRNet}
After the training and optimisation phases of the domain component, the resulting network can be used 
 to solve any goal recognition instance in the domain through the instance-specific component of our system (right part of Figure \ref{fig:architecture}).
Such component performs an evaluation of the candidate goals in $\cal G$ of the GR instance, using the output of the domain component fed by the observations of the GR instance.
To choose the most probable goal in $\mathcal{G}$ (solving the multi-class classification task associated to the GR instance), we designed a simple score function that indicates how likely it is that $G$ is the correct goal, according to the neural network of the domain component.
This score is defined as
$    score(G, \overline{o}) = \sum_{f \in G} \overline{o}_f
$,
where $\overline{o}$ is the network output vector, and  $\overline{o}_f$ is the network output for fact $f$.
For each candidate goal $G \in \cal G$, we consider only the output neurons that have associated facts in $G$, ignoring the others. By summing only these predicted values we can derive an overall score for $G$ being the correct goal.
We compute this score for all candidate goals in $\cal G$, and we select the one with the highest score ($G_1$ in Figure \ref{fig:architecture}).

\begin{example}
In our running example, since we are assuming that GR instance involve at most 22 blocks, we have that the action set $A$ is formed by 
$22$ {\small \tt Pick-Up} actions,  22 {\small \tt Put-Down} actions, $22 * 21 = 462$ {\small \tt Stack} actions and  462 {\small \tt Unstack} actions, for a total of  $968 = |A|$ different actions in the domain.
Suppose that the three observed actions 
{\small \tt (Pick-Up Block\_C)}, {\small \tt (Stack Block\_C Block\_B)} and {\small \tt (Pick-Up Block\_F)}
forming the observation sequence $O$ 
have ids corresponding to indices $5$, $17$ and $21$, respectively.
In the Domain Component of GRNet, after being processed by the embedding layer, the input $O$ is represented by the sequence of vectors $e_{05}$, $e_{17}$ and $e_{21}$. 
Then this sequence is fed to the LSTM layer and subsequently to the attention mechanism, producing a context vector $c$ representing the entire plan trace formed
by the observed actions. Finally, vector $c$ is
processed by a final feed-forward layer made of 
$\abs{F} = 506$ neurons.
In this representation, each neuron corresponds to a distinct proposition in $F$. Therefore, if the network has to predict candidate goal $G_1$, made by 
{\small \tt (On~Block\_C~Block\_B)} and {\small \tt (On~Block\,\,F~Block\_C)},
their corresponding neuron should have value 1, while the  neurons of the different propositions in $G_2$ should 
have value zero.
Our GR instance is made by the plan trace {\small \tt (Pick-Up Block\_C)}, {\small \tt (Stack Block\_C Block\_B)} and {\small \tt (Pick-Up Block\_F)} and by two possible goals: $G_1$, made by {\small \tt (On Block\_C Block\_B)} and {\small \tt (On Block\_F Block\_C)}, and $G_2$ made by  {\small \tt  (On Block\_G Block\_H)} and {\small \tt (On Block\_H Block\_F)}.
Therefore, in the Instance Component of GRNet, we calculate the prediction values of $G_1$ and $G_2$ as the sum of the predictions for the neurons representing their facts.
Suppose that $\overline{o}_{{\tt (On~Block\_C~Block\_B)}}=1.000$, $\overline{o}_{{\tt  (On~Block\_F~Block\_C)}}=0.017$, $\overline{o}_{{\tt   (On~Block\_G~Block\_H)}}=0.000$, $\overline{o}_{{\tt (On~Block\_H~ Block\_F)}}=0.003$, we have that the final score of $G_1$ is $1.017$, while the final score of $G_2$ is $0.003$.
The goal with the highest score ($G_1$) is selected as the most probable goal solving the GR instance.
\end{example}

\section{Experimental Analysis}

After describing the benchmark domains, datasets, and GR instances, we
analyse the performance of \our, comparing it with a state-of-the-art goal recognition system \cite{pereira2020landmark}.

\subsection{Benchmark Suite and Data Sets}
\label{sec:setup}
\begin{table}
\scriptsize
    \centering
    \begin{tabular}{ lccccc }
        
        \toprule
        Domain & $\abs{A}$ & $\abs{F}$ & $\abs{G_i}$ & $\abs{\mathcal{G}}$ \\
        \midrule
        \blocksworld  & 968 & 506  &   [4,16] & [19,21]\\
        \depots       & 13050 & 150 &   [2,8]  & [7,10]\\
        \driverlog    & 4860 & 156 &   [4,11] & [6,10]\\
        \logistics    & 15154 &  154 & [2,4]  & [10,12]\\
        \satellite    & 33225 & 629  & [4,9]  & [6,8]\\
        \zenotravel   & 23724 & 66  &  [5,9]  & [6,11]\\
        \bottomrule
    \end{tabular}
    \caption{Size of $A$, $F$, $G_i \in \cal G$ and $\cal G$ in the considered GR instances for each considered domain.  Interval $[x,y]$ indicates a range of integer values from $x$ to $y$.}
    \label{tab:experiments_detail}
\end{table}
 We  consider six benchmark domains that are well known in the automated planning community: \blocksworld, \depots, \driverlog, \logistics, \satellite and \zenotravel \cite{ipcfirst,ipcthird}. 

In order to create the (solved) GR instances for the training and test sets in the considered domains, we used automated planning techniques.  
Concerning the training set, 
for each domain, we randomly generated a large collection of (solvable) plan generation problems
of different size.
We considered the same ranges of the numbers of involved objects as in the experiments of  \citet{pereira2020landmark}.
For each of these problems, we computed up to four (sub-optimal) plans solving them. As planner we used  \lpg \cite{lpg_jair}, which allows to specify the number of requested different solutions for the planning problem it solves.
From the generated plans we derived the observation sequences for the training samples by randomly selecting actions from the plans (preserving their relative order). The selected actions are between 
30\% and 70\% of the plan actions.
The generated training set consists of pairs $(O, G^*)$ where $O$ is a sequence of observed actions obtained by sampling a plan $\pi$, and $G^*$ is the hidden goal corresponding to the goal of the planning problem solved by $\pi$.
For each considered domain, we created a training set with $55000$ pairs.

For evaluating \our we generated two test sets formed by GR instances {\em not seen at training time}: \ourTEST and \recogTEST.
Such test instances were generated as for the train instances, except that the observation sequences were derived using also plans generated by a planner different from the one used for creating the train instances (\lpg). As second planner we used \lama \cite{richter:10:lama}.

 \ourTEST is a generalisation and extension of the test set used in \cite{pereira2020landmark} for the same domains that we consider. We indicate this original test set with \pereiraTEST. \ourTEST includes all 
 \pereiraTEST instances, and the goal sets ($\mathcal{G}$) of \pereiraTEST and \ourTEST are the same also for the new instances.
The additional instances in \ourTEST are motivated by the limited number and particular structure of those in \pereiraTEST. 
The structure is limited because 
the observations in the instances of \pereiraTEST are created from plans for the goals in $\mathcal{G}$ that are all generated from the {\em same} initial state. In \ourTEST, the GR instance are created  considering multiple plans generated from different initial states, obtaining a richer diversification of the observation traces and a larger number of test instances. 
 In particular, for each of \depots, \driverlog, \satellite and \zenotravel, \pereiraTEST contains only $84$ instances, while \ourTEST test set contains roughly $6000$ instances.

For each plan generated for being sampled, we randomly derived three different action traces formed by 30\%,  50\% and 70\% of the plan actions, respectively. This gives three groups of test instances, for each considered domain, allowing to evaluate the performance of \our also in terms of different amounts of available observations. 
Table \ref{tab:experiments_detail}  gives information about the size of the GR instances in our test and training sets for each domain, in terms of number of possible actions ($\abs{A}$), facts ($\abs{F}$),  min/max size of a goal ($\abs{G_i}$) in a goal set $\cal G$, and min/max size of a goal set ($\abs{\mathcal{G}}$). More details in the supplementary material.

Test set \recogTEST was created to evaluate 
how well the compared systems behave on GR instances of different difficulty. We focus this analysis on a specific domain (\zenotravel).
In \recogTEST, the generated GR instances are grouped into several classes according to their difficulty. As difficulty measure 
we used the notion of \emph{recognizability of the hidden goal}, which is inspired by the notion of the ``uniqueness of landmarks" introduced by \citet{pereira2020landmark}.
Specifically, the recognizability $R(G)$ of a goal $G \in \cal G$ is defined as

$R(G) = \sum_{f \in G}{\frac{1}{|\{G' \; \mid \; G' \in \mathcal{G} \land f \in G'\}|}}$.

\noindent
The lower $R(G)$ is, the more difficult recognising $G$ is; vice versa, the higher $R(G)$ is, the more discernible $G$ is. 

We normalized $R_Z(G)$ as a value between $0$ and $1$, denoted $R_Z(G)$.
For example, if $\mathcal{G} = \set{G_1, G_2, G_3 }$, with $G^* = G_1 = \set{a,b,c}$, $G_2 = \set{ a,e,f }$ and $G_3 = \set{g,h,i}$, then $R(G^*)= \frac{1}{2} + 1 +1 = \frac{5}{2}$ and $R_Z(G^*) = 0.75$ (high recognizability). While, if 
$\mathcal{G} = \set{G_1, G_2, G_3}$ with $G^* =  G_1 = \set{a,b,c}$, $G_2 = \set{a, b, x}$ and $G_3 = \set{a,b,y}$, then $R(G^*)=\frac{1}{3} + \frac{1}{3} + 1 = \frac{5}{3}$, and so $R_Z(G^*) = 0.33$ (low recognizability). 

Using different values for $R_Z(G^*)$, we have generated nine classes of GR instances, denoted $C_1,...,C_9$. For each GR instance in class $C_i$, we have $0.1 \cdot i \leq R_Z(G^*) < 0.1 \cdot (i+1)$, for $i = 1...9$. Each class consists of 100 GR instances. 

\subsection{Experimental Results}
\label{sec:results}

We present the results obtained by \our and the state-of-the-art system \pereiraGR by \citet{pereira2020landmark}  over the test sets \ourTEST and \recogTEST. In order to have a fair comparison, for \pereiraGR we used the $h_{\textit{uniq}}$  heuristic for the goal selection, because the authors stated that it performs better than others. In \pereiraGR it is possible to set a threshold $0\leq \theta \leq 1$ that controls the confidence to accept a goal as a possible solution for the given GR instance. In order to select a single goal in $\cal G$, we set $\theta=0$. Nevertheless, we observed that in some GR tasks \pereiraGR returns more than one goal. %
As done in \cite{pereira2020landmark}, for \pereiraGR  we considered solved a GR instance when their system returns a set of goals {\em containing} the correct goal (making the evaluation more favourable for \pereiraGR).

Goal recognition {\em accuracy} for a set of test instances is defined as the percentage of instances whose goals are correctly identified (predicted) over the total number of the instances in the set. 

\begin{table}[]
    \centering
    \scriptsize
    \begin{tabular}{ l|cc|cc|cc }
        \toprule
        \multirow{2}{*}{Domain} & \multicolumn{2}{c |}{30\% of  the plan} & \multicolumn{2}{c|}{50\% of the plan} & \multicolumn{2}{c}{70\% of the plan} \\
        & \pereiraGR & \our & \pereiraGR & \our & \pereiraGR & \our\\
        \midrule
        \blocksworld & 38.9 & \bf{53.4*} & 52.9 & \bf{69.4*} & 76.0 & \bf{82.6} \\
        \depots & 45.9 & \bf{60.9*} & 65.4 & \bf{75.8*} & 82.1 & \bf{86.3} \\
        \driverlog & 43.9 & \bf{61.4*} & 61.5 & \bf{74.8*} & 79.7 & \bf{83.9} \\
        \logistics & 50.1 & \bf{65.7*} & 67.8 & \bf{78.5*} & 82.6 & \bf{87.4} \\
        \satellite & 69.0 & \bf{71.9} & 83.2 & \bf{84.0} & 90.3 & \bf{93.7} \\
        \zenotravel & 47.9 & \bf{77.0*} & 68.8 & \bf{89.7*} & 87.1 & \bf{96.0} \\
        \bottomrule
    \end{tabular}
    \caption{Goal classification accuracy (percentages of GR instances correctly solved) of \pereiraGR and \our over six domains. Bold results indicate better performance;
    ``*" indicates accuracy improvements of \our wrt \pereiraGR of at least $10$ points.
    }
    \label{tab:results}
\end{table}

\paragraph{Results for \ourTEST.}
Table \ref{tab:results} summarizes the performance results of \pereiraGR and \our in terms of accuracy with test set \ourTEST. 
We focus the analysis on test instances derived from a mix of plans generated by \lama  and \lpg (half of the instances from each of the planners, for every domain and percentage of observed actions). It should be noted that very similar performance results were obtained using plans from either only \lpg or only \lama; these results are in the supplementary material.

\our performs generally well.
Even for instances with only $30\%$ of the actions, it shows interesting performance, with the best result for \zenotravel where \our reaches accuracy $77$. 
With 50\% of the actions, the accuracy of \our improves in every domain by more than $10$ points. For instance, in $\satellite$ \our improves from $72.8$ to $84.0$. With 70\% of the actions, our system improves further reaching accuracy above $80$ in all domains, and obtaining an impressive performance for \zenotravel ($95.9$). 
Compared to \pereiraGR, \our performs always better, for every considered domain and percentage of observations.
In many cases, the performance improvement wrt \pereiraGR is at least of $10$ points.

Moreover, \our's performance does not seem to be affected by the diversity of the domains indicated by the four parameters of Table \ref{tab:experiments_detail}.
While the remarkable performance obtained for \zenotravel might be correlated with the fact that in this domain the test instances have only $66$ facts (see column $\abs{F}$ of Table \ref{tab:experiments_detail}),  the results for \satellite are not so distant even if the instances in this domain have $629$ facts. Analysing the experimental results, it seems that also the number of the actions has no significant impact on the performance. In fact, while \blocksworld has only $968$ actions, the other domains have more than $15000$ actions, and \our obtain better results for them. This is probably due to the embedding layer that is able to learn a compact and informative representation even with a large vocabulary of actions. Overall, \our exhibits good robustness with respect to the size of the space of actions and the number of facts in the domains (the output of the network).

In term of CPU time required to solve (classify) a GR instance, \our performs generally much better than \pereiraGR. The average execution time of \pereiraGR is $1.158$ seconds with a standard deviation of $0.87$ seconds, while \our runs on average in $0.06$ seconds with a standard deviation of $0.04$ seconds. For lack of space, details about CPU times for each domain are given in the Supplementary Material.

While we consider the evaluation of \our using \ourTEST more significant and more informative than using  \pereiraTEST, we performed a comparison of 
\our and \pereiraGR also using \pereiraTEST, obtaining results that overall are in favor of \our also for this restricted test set.

\begin{figure}[t]
    \centering
    \includegraphics[width=0.90\linewidth]{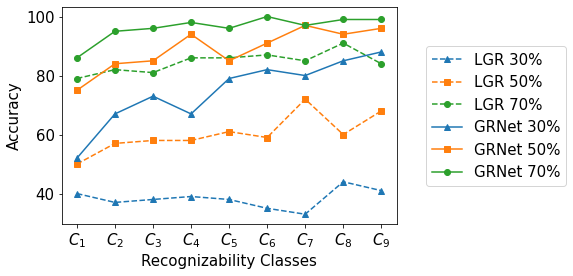}
    \caption{
    Accuracy results of \pereiraGR and \our  on several GR instances grouped into classes of decreasing difficulty with test set \recogTEST. $C_1$ is the most difficult class while $C_9$ is the easiest one.}
    \label{fig:uniq_exp}
\end{figure}

\paragraph{Results for \recogTEST.} 
Figure \ref{fig:uniq_exp} shows the accuracy of the two compared systems considering different classes of test sets with decreasing difficulty measured using $R_Z$. As expected, the accuracy of \our depends on the difficulty of the problem, since there is an increasing trend in terms of accuracy for each observation percentage. This trend is more evident when we have $30\%$ of the actions and becomes less marked as the number of observations grows. \pereiraGR appears to be more stable over the recognizability classes than \our. However, \our always performs significatly better than \pereiraGR regardless the value of $R_Z$.

\begin{figure}
    \centering
    \includegraphics[width=0.64\linewidth]{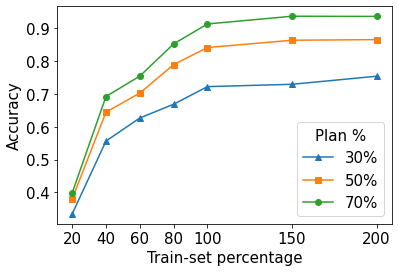} 
    \caption{Accuracy of \our trained using data sets of GR examples with different sizes (percentages of the original train set) using test set \ourTEST
    in domain \satellite.}
    \label{fig:incr_training_our}
\end{figure}

\paragraph{Sensitiveness to the training set size.} 
Since the predictive performance of a machine learning system can be deeply influenced by the number of training instances, we experimentally investigated how much \our is sensible to the this issue. We focused this analysis on domain \satellite, training several neural networks with different fractions of our training set: 20\%, 40\%, 60\% and 80\%. Figure \ref{fig:incr_training_our} shows how accuracy increases for
\ourTEST
when the training set size increases. In particular, for \ourTEST we can observe that using only 20\% of the training instances gives accuracy lower than 40 in all three cases (30-50-70\% of observed actions), but accuracy rapidly improves reaching more than 60 using 60\% of the training instances.

We evaluated \our also for larger training sets, up to twice the number of instances in the original training set. 
As it can be seen in Figure \ref{fig:incr_training_our},
the enlarged training set for
\ourTEST produces only a small improvement in accuracy.

\section{Related Work}
Goal recognition and plan recognition have been extensively studied through model-based approaches exploiting planning techniques \cite{DBLP:conf/ijcai/MeneguzziP21,geffner,gr_noisy,santos,pereira2020landmark} or matching techniques relying on plan libraries (e.g., \cite{plan_libraries, mirsky}).
We have presented a model-free approach, 
discussing its advantages and experimentally comparing it with the state-of-the-art model-based approach \pereiraGR \cite{pereira2020landmark}. 
This is a planning-based system exploiting landmarks \cite{hoffmann2004ordered}.
Differently from \our, \pereiraGR
 requires domain knowledge and performs no learning from previous experiences. Similarly to \our , the output goal provided by \pereiraGR is not guaranteed to be correct.

Many approaches to human activity recognition adopt architectures similar to the RNN of \our \cite{9760794, yin2022novel, chen2016lstm, khatun2022deep}. These systems, though, address a problem different from goal recognition; they deal with noisy input data from sensors, and perform a specific classification task (with fixed classification values). As a consequence, these architectures provide solutions to very specific problems. The work in this paper addresses the goal recognition problem, it deals with lack of observability in the actions of the agent's plan, and proposes a more general approach allowing to solve, by the same trained network, different goal recognition instances in the domain.

Concerning GR systems using neural networks, some works use them 
for specific applications, such as game playing \cite{min}. \our is more general, as it can be applied to any domain of which we know sets $F$ and $A$.
In order to extract useful information from image-based domains and perform goal recognition, \citet{amado_images} used a pre-trained encoder and a LSTM network for representing and analysing sequence of observed states, rather than actions as in our approach.
\citet{amado_observations} trained a LSTM-based system to identify missing observations about states in order to derive a more complete sequence of 
states by which a MBGR system can obtain better performance. 

\citet{borrajo} investigated the use of XGBoost \cite{chen2016xgboost} and LSTM neural networks for goal recognition using
only traces of plans, without any knowledge on the domain, similarly to  our approach. 
However, \citeauthor{borrajo} train a specific machine learning model for each goal recognition instance (the goal set $\cal G$ is fixed), using instance-specific datasets for training and testing. Instead, in our approach we train a general-purpose neural network that can be used to solve a large number of different goal recognition instances, without the need of designing or training a new model. 
Moreover, the experimental evaluation of the networks proposed in \cite{borrajo}
use peculiar goal recognition benchmarks with custom-made instances.
Instead, in our work we evaluate \our much more in depth using known benchmarks introduced by \cite{pereira2020landmark}, an extension of them having many more testing instances, and additional benchmarks based with different degrees of goal recognition difficulty.

\citet{maynard} compared model-based techniques and approaches based on deep learning for goal recognition. However, as in \cite{borrajo}, such a comparison is made using specific instances, and several kinds of neural networks are trained to
predict directly the goal among a set of possible ones, instead of the facts that belong to the goal as in our approach. This makes the trained networks in \cite{maynard} specific for the considered GR instances in a domain, while our approach is more general since it trains a single network for the  domain. 

Another substantial difference is that, while in a typical goal recognition problem we can have missing observation across the entire plan of the agent(s), the work in \cite{maynard} considers only observations from the start of the plan to a given percentage of it, treating every possible successive observation as missing.

\citet{amado2022goal} proposed a framework that combines off-the-shelf model-free reinforcement learning and state-of-the-art goal recognition techniques achieving promising results. However, similarly to \cite{borrajo}, their approach is designed to solve a specific goal recognition instance where the goal set $\mathcal{G}$ is fixed. On the contrary, the trained RNN of \our can  be used to solve all GR instances definable over the domain and action sets ($F$ and $A$).

\section{Conclusions}

We have proposed an approach to address goal recognition as a deep learning task. Our system, called \our, learns to solve (classify) goal recognition tasks from past experience in a given domain, and
requires no model of the domain actions nor a specification of an initial state.
Learning consists in training only one neural network for the considered domain, allowing to solve a large collection of GR instances in the domain by same trained network.
An experimental analysis shows that \our performs well in several benchmark domains,
in terms of both accuracy and runtime of the trained system, outperforming a state-of-the-art GR system.

The GR tasks addressable by \our are limited to those involving subsets of  facts and actions appearing in the training set. An interesting question for future work is how to extend \our to solve GR instances involving new actions and facts.
We also intend to (i) carry out more experiments to evaluate \our and
investigate how its RNN can rich high performance (e.g., by analysing the weights provided by the attention mechanism and the learning process), and (ii) study the use of the more recent deep learning architectures.

Finally, an interesting direction for future work concerns finding effective ways of combining the model-free and the model-based approaches, the first based on learning from past experience and the second exploiting automated reasoning.
While a loose combination is using planning to generate samples in the training dataset for a learning-based GR system, finding a tighter integration that can further improve goal recognition performances is for future research.

\newpage
\bibliography{bibliography.bib}

\newpage

\onecolumn
\section{Supplementary Material}
\label{sec:supplementary-data}

\subsection{Domains description}
We provide a very brief description of the six well-known planning domains used in our experiments:
\begin{itemize}
\item \textbf{\blocksworld}. The domain consists of an hand robot that has to stack or unstack blocks, picking up them one at a time, in order to obtain a desired configuration of an available set of blocks.

\item \textbf{\depots}. The domain consists of actions for loading and unloading packages into trucks through hoists, and moving them between depots. The goals concern having the packages at certain depots.

\item \textbf{\driverlog}. In this domain there are drivers (trucks) that can walk (drive) between locations. Walking and driving require traversing different paths. Packages can be loaded into or unloaded from trucks, that can be moved by drivers. The goal is to deliver (move) all packages (drivers) to their destinations. 

\item \textbf{\logistics}. In this domain there are  aircrafts that can fly between cities, trucks that can move between locations within a city, and packages that can be loaded into/unloaded from trucks and aircrafts. The goal is to deliver a set of packages to their delivery locations.

\item \textbf{\satellite}. This is a scheduling domain in which one or more satellites can make certain observations, collect data, and download the collected data to a ground station. The goals concern having observation data at a ground station.

\item \textbf{\zenotravel}. This is another variant of a  transportation domain where passengers have to be embarked and disembarked into aircrafts, that can fly between cities at two possible speeds. The goals concern transporting (move) all passengers (aircrafts) to their required destinations.
\end{itemize}

\subsection{Size of the GR instances for the training/test sets in the considered benchmark domains}\label{ranges}
We give details about the number of objects involved in the GR instances.  
For each object type in a domain, we report its name, the minimum and the maximum number of objects of that type ($min$ and $max$) that are involved in an instance, and the total number of objects of that type in the domain ($objs$); $objs$ indicates the number of all possible objects of a type that can be used to define a GR instance using at most $max$ of them. We chose these $min$-$max$ ranges because they are the {\em same} as those used in the experiments of \citet{pereira2020landmark}.

\begin{itemize}
    \item \blocksworld. \set{
    block: \set{min: 7, max: 17, objs: 22}
    }
        
    \item \depots, \set{
    depot: \set{min: 1, max: 3, objs: 3 },\;
    distributor: \set{min: 1, max: 3, objs: 3},\;
    truck: \set{min: 2, max: 3, objs: 3 },\;
    pallet: \set{min: 2, max: 6, objs: 6},\;
    crate: \set{min: 2, max: 10, objs: 10},\;
    hoist: \set{min:2, max: 6, objs: 6 }
    }
    
    \item \driverlog, \set{
    driver: \set{min: 2, max: 3, objs: 3},\;
    truck: \set{min: 2, max: 3, objs: 3},\;
    package: \set{min: 2, max: 7, objs: 7},\;
    locations_s: \set{min: 3, max: 12, objs: 12},\; 
    location_p: \set{min: 2, max: 25, objs: 41}
    }
    
    \item \logistics, \set{
    airplane: \set{min: 1, max: 8, objs: 8},\;
    airport: \set{min: 2, max: 8, objs: 8},\;
    location: \set{min: 6, max: 11, objs: 11},\;
    city: \set{min: 2, max: 6, objs: 6},\;
    truck: \set{min: 2, max: 5, objs: 5},\;
    package: \set{min: 2, max: 14, objs: 14}
    }
    
    \item \satellite, \set{
    satellite: \set{min: 1, max: 5, objs: 5},\;
    instrument: \set{min: 1, max: 11, objs: 11},\;
    mode: \set{min: 3, max: 5, objs: 12},\;
    direction: \set{min: 7, max: 17, objs: 37}}

    \item \zenotravel, \set{
    aircraft: \set{min: 2, max: 3, objs: 3},\;
    person: \set{min: 5, max: 8, objs: 8},\;
    city: \set{min: 3, max: 6, objs: 6},\;
    flevel: \set{min: 7, max: 7, objs: 7}
    }

\end{itemize}

\subsection{Hyperparameters of the Neural Network}

Table \ref{tab:hyperparameters_range_sup} reports the hyperparameters range used in the Optuna framework. For each benchmark domain, we used a separate process of optimization (study) which executes 30 objective function evaluations (trials). We used a sampler that implements the Tree-structured Parzen Estimator algorithm.

\vspace{0.2cm}
\noindent
Table \ref{tab:hyperparameters_sup} reports the hyperparameters of the neural networks in our experiments. For all the experiments we selected 64 as batch size; we used Adam as optimizer with $\beta_1 = 0.9 $ and $\beta_2 = 0.99$.

\newpage

\begin{table}[h!]
\centering

\begin{tabular}{lc}
\toprule
Hyperparameter & Range\\
\midrule
${\abs{E}}$ & [50, 200]\\
${\abs{LSTM}}$ & [150, 512]\\
Use Dropout & \{True, False\}\\
Dropout & [0, 0.5]\\ 
Use Rec. Dropout & \{True, False\}\\
Rec. Dropout & [0, 0.5]\\

\bottomrule
\end{tabular}
\caption{
Value ranges of the hyperparameters for the Bayesian-optimisation done by the Optuna framework. ${\abs{E}}$ represents the dimension of the embedding vectors, ${\abs{LSTM}}$ is the number of neurons in the LSTM layer.  Interval [x, y] indicates a range of integer values from x to y, while set $\{x_1,..x_n\}$ enumerates all possible values the hyperparameter can assume.}
\label{tab:hyperparameters_range_sup}
\end{table}

\begin{table}[h!]
\centering

\begin{tabular}{lcccc}
\toprule
Domain & ${\abs{E}}$ & ${\abs{LSTM}}$ & Dropout & Recurrent Dropout\\ 
\midrule
\blocksworld & 119 & 354 & 0.00 & 0.00\\ 
\depots      & 200 & 450 & 0.15 & 0.23 \\ 
\driverlog   & 183 & 473 & 0.00 & 0.00 \\ 
\logistics   & 85 & 446 & 0.12 & 0.01 \\ 
\satellite   & 117 & 496 & 0.04 & 0.00  \\ 
\zenotravel  & 83 & 350 & 0.00 & 0.00 \\ 
\bottomrule
\end{tabular}
\caption{Hyperparameters of the networks used in our experiments. ${\abs{E}}$ is the size of the vector in output from the embedding layer, ${\abs{LSTM}}$ is the number of neurons in the LSTM layer.}
\label{tab:hyperparameters_sup}
\end{table}

\subsection{Test Set Instances}

Table \ref{tab:test_support_a}  reports the number of GR instances for the test sets used in our experiments.
Domains \blocksworld and \satellite have fewer instances 
 to avoid the overlapping between train and test instances, because of the more limited  space of GR instances in these domains.

\begin{table}[h!]
\centering
\begin{tabular}{lccc}
\toprule
Domain & $\abs{\pereiraTEST}$ & $\abs{\ourTEST} $ & $\abs{\recogTEST}$ \\ 
\midrule
\blocksworld & 246 & 3136 & 900 \\ 
\depots      & 84 & 6720 & 900 \\ 
\driverlog   & 84 & 6720 & 900 \\ 
\logistics   & 153 & 6720 & 900 \\ 
\satellite   & 84 & 5760 & 900 \\ 
\zenotravel  & 84 & 6720 & 900 \\ 
\bottomrule
\end{tabular}
\caption{Number of GR instances per domain for the test sets used in the experiments.}
\label{tab:test_support_a}
\end{table}

\subsection{Additional Material about the Experimental Analysis}\label{detailed_results}

Table \ref{tab:results_extra_our} reports results about the performance of \our using \ourTEST in terms of Accuracy, Precision, Recall and F-Score.

\vspace{0.2cm}
\noindent
Table \ref{tab:results_extra_per} reports results about the performance of \pereiraGR using \ourTEST in terms of Accuracy, Precision, Recall and F-Score.

\vspace{0.2cm}
\noindent
Table \ref{tab:results_lama_lpg_appendix} reports results about the performance of \our in terms of accuracy with the instances in \ourTEST created from plans generated using either \lpg, \lama, or both planners. 

\vspace{0.2cm}
\noindent
Table \ref{tab:times} shows the average execution times (milliseconds) of\pereiraGR and \our for tests set \ourTEST. 

\begin{table}[h!]
\small
    \centering
    \begin{tabular}{ lcccccc }
        \toprule
        Domain & Observations & Plan Length & Accuracy  & Precision & Recall & F1-Score \\
        \midrule
        \multirow{3}{*}{\blocksworld} & 30\% & 10.1 & 53.4 & 54.4 & 53.2 & 52.1\\
        & 50\% & 17.5 & 69.4 & 71.2 & 70.2 & 69.0\\
        & 70\% & 24.1 & 82.6 & 83.2 & 81.8 & 81.1\\
        \hline
        \multirow{3}{*}{\depots} & 30\% & 8.4 & 60.9 & 60.9 & 60.8 & 60.6 \\
        & 50\% & 14.6 & 75.8 & 76.6 & 76.4 & 76.3\\
        & 70\% & 20.3 & 86.3 & 87.3 & 87.1 & 87.1\\
        \hline
        \multirow{3}{*}{\driverlog} & 30\% & 7.0 & 61.4 & 61.7 & 61.9 & 61.4 \\
        & 50\% & 12.2 & 74.8 & 75.9 & 74.1 & 75.0\\
        & 70\% & 17.0 & 83.9 & 84.7 & 84.2 & 83.1\\
        \hline
        \multirow{3}{*}{\logistics} & 30\% & 7.4 & 65.7 & 67.9 & 66.8 & 66.0 \\
        & 50\% & 12.9 & 78.5 & 79.4 & 78.5 & 78.0 \\
        & 70\% & 17.9 & 87.4 & 87.3 & 86.7 & 86.3\\
        \hline
        \multirow{3}{*}{\satellite} & 30\% & 4.7 & 71.9 & 72.2 & 72.3 & 70.9 \\
        & 50\% & 8.3 & 84.0 & 83.2 & 83.3 & 82.2 \\
        & 70\% & 11.6 & 93.7 & 92.0 & 93.0 & 92.2 \\
        \hline
        \multirow{3}{*}{\zenotravel} & 30\% & 6.1  & 77.0 & 77.5 & 76.6 & 76.5\\
        & 50\% & 10.6 & 89.7 & 90.0 & 89.5 & 89.5\\
        & 70\% & 14.8 & 96.0 & 96.0 & 95.9 & 95.8\\
        \bottomrule
    \end{tabular}
    \caption{Experimental results about the performance of \our. 
    Columns Observations and Plan Length correspond  to the observability percentage over the full plan and the average number of observed actions in test instances, respectively.
    Each value in column Precision is an average of the precision for the gropus of instances that have the same goal set $\cal G$. Similarly for the values of columns Recall and F1 score. 
    } 
    
    \label{tab:results_extra_our}
\end{table}

\newpage
\begin{table}[h!]
    \small
    \centering
    \begin{tabular}{ lcccccc }
        \toprule
        Domain & Observations & Plan Length & Accuracy  & Precision & Recall & F1-Score \\
        \midrule
        \multirow{3}{*}{\blocksworld} & 30\% & 10.1 & 38.9 & 39.3 & 37.6 & 37.3\\
        & 50\% & 17.5 & 52.9 & 59.6 & 52.4 & 53.0\\
        & 70\% & 24.1 & 76.0 & 79.3 & 75.6 & 76.0\\
        \hline
        \multirow{3}{*}{\depots} & 30\% & 8.4 & 45.9 & 46.4 & 45.9 & 45.7 \\
        & 50\% & 14.6 & 65.4 & 65.9 & 65.0 & 64.9\\
        & 70\% & 20.3 & 82.1 & 82.3 & 82.1 & 82.1\\
        \hline
        \multirow{3}{*}{\driverlog} & 30\% & 7.0 & 43.9 & 45.3 & 43.9 & 44.0\\
        & 50\% & 12.2 & 61.5 & 62.7 & 61.6 & 61.7\\
        & 70\% & 17.0 & 79.7 & 80.1 & 79.6 & 79.7\\
        \hline
        \multirow{3}{*}{\logistics} & 30\% & 7.4 & 50.1 & 51.1 & 50.6 & 50.6 \\
        & 50\% & 12.9 & 67.8 & 68.7 & 68.3 & 68.1 \\
        & 70\% & 17.9 & 82.6 & 83.7 & 83.7 & 83.4\\
        \hline
        \multirow{3}{*}{\satellite} & 30\% & 4.7 & 69.0 & 71.5 & 68.8 & 68.0 \\
        & 50\% & 8.3 & 83.2 & 83.0 & 82.3 & 81.5 \\
        & 70\% & 11.6 & 90.3 & 89.5 & 90.0 & 89.2 \\
        \hline
        \multirow{3}{*}{\zenotravel} & 30\% & 6.1  & 47.9 & 48.0 & 47.6 & 47.7\\
        & 50\% & 10.6 & 68.8 & 68.9 & 68.9 & 68.9\\
        & 70\% & 14.8 & 87.1 & 87.1 & 87.1 & 87.1\\
        \bottomrule
    \end{tabular}
    \caption{
    Experimental results about the performance of \pereiraGR.
    Columns Observations and Plan Length correspond  to the observability percentage over the full plan and the average number of observed actions in test instances, respectively.
    Each value in column Precision is an average of the precision for the gropus of instances that have the same goal set $\cal G$. Similarly for the values of columns Recall and F1 score. 
    }
    \label{tab:results_extra_per}
\end{table}

\begin{table}[]
    \centering
    \begin{tabular}{ l|ccc|ccc|ccc }
        \toprule
        \multirow{2}{*}{Domain} & \multicolumn{3}{c |}{30\% of  the plan} & \multicolumn{3}{c|}{50\% of the plan} & \multicolumn{3}{c}{70\% of the plan} \\
        & \lpg & \lama & \lpg/\lama & \lpg & \lama & \lpg/\lama & \lpg & \lama & \lpg/\lama \\
        \midrule
        \blocksworld & 52.4 & 54.4 & 53.4 & 70.0 & 68.8 & 69.4 & 82.3 & 82.9 & 82.6 \\
        \depots & 61.0 & 60.9 & 60.9 & 76.5 & 75.2 & 75.8 & 87.2 & 85.5 & 86.3 \\
        \driverlog & 64.9 & 57.9 & 61.4 & 78.1 & 71.5 & 74.8 & 86.2 & 81.6 & 83.9 \\
        \logistics & 66.8 & 64.6 & 65.7 & 78.5 & 78.4 & 78.5 & 86.8 & 88.0 & 87.4 \\
        \satellite & 72.7 & 71.2 & 71.9 & 84.0 & 83.9 & 84.0 & 93.6 & 93.9 & 93.7 \\
        \zenotravel & 77.6 & 76.4 & 77.0 & 89.4 & 89.9 & 89.7 & 95.9 & 96.1 & 96.0 \\
        \bottomrule
    \end{tabular}
    \caption{Goal classification accuracy of \our 
    for  \ourTEST instances created from plans generated by \lpg, \lama, and both planners over six domains.
    }
    \label{tab:results_lama_lpg_appendix}
\end{table}

\begin{table}[h!]
    \centering
    \begin{tabular}{ l|cc|cc|cc|cc}
        \toprule
        \multirow{2}{*}{Domain} & \multicolumn{2}{c |}{30\%} & \multicolumn{2}{c|}{50\%} & \multicolumn{2}{c |}{70\%} & \multicolumn{2}{c}{Overall mean} \\
        & \pereiraGR & \our & \pereiraGR & \our & \pereiraGR & \our & \pereiraGR & \our \\
        \midrule
        \blocksworld & 832 & 51 & 830 & 50 & 839 & 51 & 834 & 51  \\
        \depots & 1138 & 65 & 1133 & 82 & 1150 & 67 & 1140 & 71 \\
        \driverlog & 1171 & 71 & 1170 & 67 & 1163 & 68 & 1168 & 69 \\
        \logistics & 1263 & 62 & 1261 & 75 & 1262 & 64 & 1262 & 67\\
        \satellite & 1507 & 64 & 1551 & 62 & 1543 & 62 & 1534 & 62\\
        \zenotravel & 1556 & 55 & 1576 & 52 & 1568 & 53 & 1567 & 54 \\
        \bottomrule
    \end{tabular}
    \caption{Comparison of the execution times of \pereiraGR and \our in milliseconds.
    }
    \label{tab:times}
\end{table}

\end{document}